\documentclass[conference]{IEEEtran}
\IEEEoverridecommandlockouts
\usepackage{cite}
\usepackage{amsmath,amssymb,amsfonts,amsthm}
\usepackage{algorithm}
\usepackage{algorithmic}
\usepackage{graphicx}
\usepackage{textcomp}
\usepackage{xcolor}
\usepackage[bookmarksnumbered=false,colorlinks=true]{hyperref}
\usepackage{bookmark}
\usepackage[all]{hypcap}
\usepackage{eso-pic}

\newcommand{\etal}{\textit{et al}.}
\newcommand{\ie}{\textit{i}.\textit{e}.}

\newcommand{\relmiddle}[1]{\mathrel{}\middle#1\mathrel{}}

\begin{document}
% IEEE Copyright notice
\AddToShipoutPicture*{\put(40,30){\sf \small \parbox[b]{12cm}{
	978-1-5386-6805-4/18/\$31.00 \textcopyright\ 2018 IEEE \\
	DOI 10.1109/ICMLA.2018.00054
}}}
\def\BibTeX{{\rm B\kern-.05em{\sc i\kern-.025em b}\kern-.08em
    T\kern-.1667em\lower.7ex\hbox{E}\kern-.125emX}}

\title{Adam Induces Implicit Weight Sparsity in Rectifier Neural Networks}
\author{
	\IEEEauthorblockN{
		Atsushi Yaguchi\IEEEauthorrefmark{1}, 
		Taiji Suzuki\IEEEauthorrefmark{2}\IEEEauthorrefmark{3}, 
		Wataru Asano\IEEEauthorrefmark{1}, 
		Shuhei Nitta\IEEEauthorrefmark{1}, 
		Yukinobu Sakata\IEEEauthorrefmark{1}, 
		Akiyuki Tanizawa\IEEEauthorrefmark{1}
	}
	\IEEEauthorblockA{
		\IEEEauthorrefmark{1}Corporate Research and Development Center, Toshiba Corporation, Kawasaki, Japan \\
		\IEEEauthorrefmark{2}Graduate School of Information Science and Technology, The University of Tokyo, Japan \\
		\IEEEauthorrefmark{3}Center for Advanced Integrated Intelligence Research, RIKEN, Tokyo, Japan
	}
	\small
	\IEEEauthorrefmark{1}\texttt{\{atsushi.yaguchi, wataru.asano, shuhei.nitta, yuki.sakata, akiyuki.tanizawa\}@toshiba.co.jp} \\
	\IEEEauthorrefmark{2}\texttt{taiji@mist.i.u-tokyo.ac.jp}
	\normalsize
}
\maketitle

%%%%%%%%%%%%%%%%%%%%%%%
%abstract
\begin{abstract}
In recent years, deep neural networks (DNNs) have been applied to various machine leaning tasks, including image recognition, speech recognition, and machine translation. 
However, large DNN models are needed to achieve state-of-the-art performance, 
exceeding the capabilities of edge devices. Model reduction is thus needed for practical use. 
In this paper, we point out that deep learning automatically induces group sparsity of weights, in which all weights connected to an output channel (node) are zero, when training DNNs under the following three conditions: (1) rectified-linear-unit (ReLU) activations, (2) an $L_2$-regularized objective function, and (3) the Adam optimizer.
Next, we analyze this behavior both theoretically and experimentally, and propose a simple model reduction method: eliminate the zero weights after training the DNN. 
In experiments on MNIST and CIFAR-10 datasets, we demonstrate the sparsity with various training setups.
Finally, we show that our method can efficiently reduce the model size and performs well relative to  methods that use a sparsity-inducing regularizer.
\end{abstract}

%%%%%%%%%%%%%%%%%%%%%%%

%%%%%%%%%%%%%%%%%%%%%%%
%introcuction
\section{Introduction}\label{sec:introduction}
Recently, deep learning has been successfully applied in various machine learning tasks \cite{LeCun_2015}. 
For example, it surpassed human performance at an image recognition task on the ImageNet dataset \cite{He_2015}. 
These successes are supported by the development of activation functions such as ReLU \cite{Glorot_2011}, 
regularization and normalization methods such as dropout \cite{Srivastava_2014} and batch normalization (BN) \cite{Ioffe_2015}, and network architectures such as ResNet \cite{He_2016}. 
Recent successes are also enabled by the growth of training datasets and increases in computing power. 
Because wider or deeper network models have become necessary to achieve high performance, 
hardware with limited memory and computational power are unable to match this performance. 
Reducing model size is necessary if applications, such as video recognition for automated driving systems and surveillance systems, are to run on edge devices, 
but this reduction of model size lowers accuracy. 

\begin{figure}[!t] 
	\centerline{\includegraphics[width = 85mm]{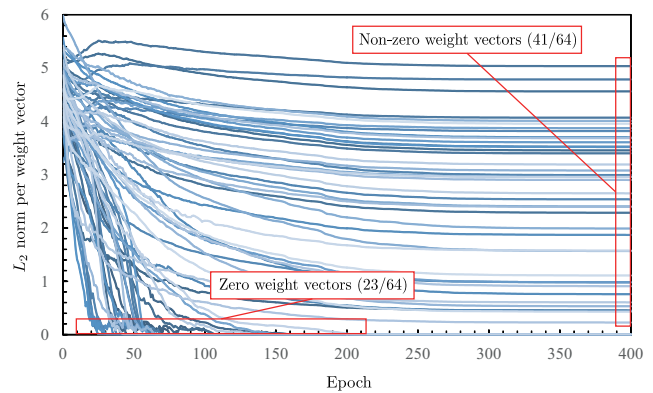}}
	\caption{
		Progress of $L_2$ norm with 64 convolution filters in first layer of a VGG-style network. 
		The network was trained with the Adam optimizer on the CIFAR-10 dataset.
	}
	\label{fig:progress_l2_norm}
\end{figure}

Various methods have been proposed for model reduction \cite{Shi_2017,Han_2016,Li_2017,Polyak_2015,Wen_2016,Scardapane_2017,Yoon_2017}. 
Shi \etal~\cite{Shi_2017} and Han \etal~\cite{Han_2016} reduced the amount of computation and data without changing the original network architecture. 
Shi \etal~\cite{Shi_2017} did this by exploiting the sparsity of ReLU activations to reduce the computational load of convolution. 
Han \etal~\cite{Han_2016}, in contrast, compressed the amount of data by quantizing weight coefficients. 
Other model reduction methods that slim the network architecture have been proposed \cite{Li_2017,Polyak_2015,Wen_2016,Scardapane_2017}. 
After training redundant models, less-important channels or nodes are pruned based on the $L_1$ norm \cite{Li_2017} or statistics of the activations \cite{Polyak_2015}. 
These methods require training redundant models before pruning. To address this, Wen \etal~\cite{Wen_2016}, Scardapane \etal~\cite{Scardapane_2017}, and Yoon \etal~\cite{Yoon_2017} proposed methods to directly train compact models. 
Their methods regularize the weights so that only some weights have nonzero coefficients, and eliminate zero weights after training to produce a sparse model. 
They reported that group sparse regularization, which induces all weights connected to an output channel (node) to be zero, 
greatly reduces the size of models because it can directly eliminate channels or nodes. 

\begin{figure*}[tb]
	\centerline{\includegraphics[width = 175mm]{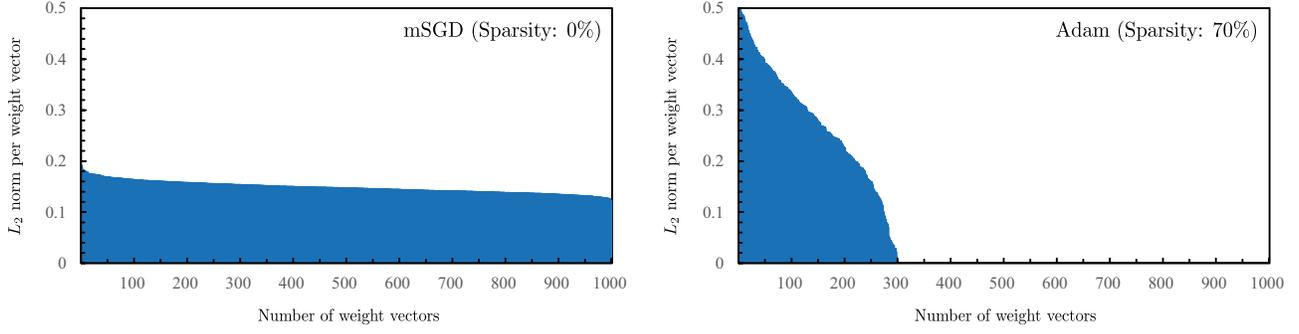}}
	\caption{
		$L_2$ norm for 1000 weight vectors connecting from an input layer to a hidden layer in a 3-layer fully connected neural network (sorted in descending order). 
		The network was trained with mSGD and the Adam optimizer on the MNIST dataset with the same settings for $L_2$ regularization and number of training steps. ``Sparsity'' indicates the ratio of zero weight vectors to all vectors.
	}
	\label{fig:adam_sparsity}
\end{figure*}

In this paper, first we point out there appears group sparsity of weights during training of DNNs under the following three conditions: (1) ReLU activations, (2) an $L_2$-regularized objective function, and (3) the Adam optimizer \cite{Kingma_2015}. 
As shown in Fig.~\ref{fig:progress_l2_norm}, though $L_2$ regularization of weights does not intrinsically induce sparsity, some weight vectors converge toward the origin (\ie, zero) if these three conditions are met. 
Interestingly, this behavior does not appear with the momentum-SGD (mSGD) optimizer \cite{Qian_1999}, as shown in Fig.~\ref{fig:adam_sparsity}. 
Next, we analyze this behavior both theoretically and experimentally, and propose a simple model reduction method: eliminate the zero weights after training the DNN. 
Our method is easy to optimize since it does not rely on sparsity-inducing (non-smooth) regularizers such as $L_{2,1}$ norm, and can efficiently control the model size by calibrating the parameter of $L_2$ regularization.
In experiments, we demonstrate the sparsity on the MNIST and CIFAR-10 datasets with various training setups, and show our method can effectively reduce the model size relative to methods that use a sparsity-inducing regularizer. 

In the following, we first give our problem setting and propose a method for model reduction (Section \ref{sec:proposed_method}), then describe its theoretical analysis (Section \ref{sec:theoretical_analysis}).
After we briefly review related works (Section \ref{sec:related_work}), we give some experimental results (Section \ref{sec:experiments}) and conclude the paper (Section \ref{sec:conclusions}).
%%%%%%%%%%%%%%%%%%%%%%%

%%%%%%%%%%%%%%%%%%%%%%%
%Proposed Method
\section{Proposed Method}\label{sec:proposed_method}

%Problem Setting
\subsection{Problem Setting}\label{subsec:problem_setting}
\begin{figure}[tb] 
	\centerline{\includegraphics[width = 85mm]{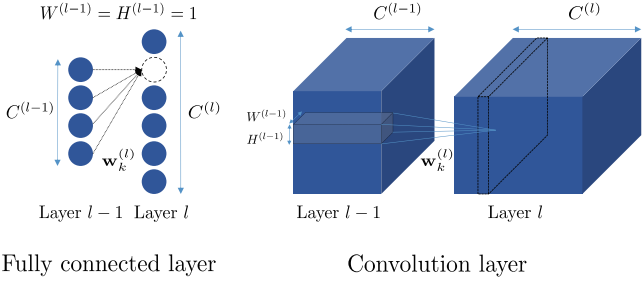}}
	\caption{Illustrations of weight vectors in fully connected and convolution layers.}
	\label{fig:weight_vectors}
\end{figure}

We consider an $L$-layer network with parameters ${\bf \Theta} = \left \{ {\bf W}^{\left( l \right)}, {\bf b}^{\left( l \right)} \right \}_{l=2}^L$, where 
$ {\bf W}^{\left ( l \right)}  = \left ( {\bf w}_1^{\left ( l \right)},{\bf w}_2^{\left ( l \right)}, \dots,{\bf w}_{C^{\left ( l \right)}}^{\left ( l \right)} \right ) \in \mathbb{R}^{W^{\left ( l-1 \right)} H^{\left ( l-1 \right)} C^{\left ( l-1 \right)} \times C^{\left ( l \right)} } $ 
is the weight matrix between layer $l-1$ and layer $l$ whose columns are the per-channel weight vectors. The corresponding bias is 
$ {\bf b}^{\left ( l \right)}  = \left ( b_1^{\left ( l \right)},b_2^{\left ( l \right)}, \dots ,b_{C^{\left ( l \right)}}^{\left ( l \right)} \right )^\top \in \mathbb{R}^{C^{\left ( l \right)} } $. 
The kernel width is $W^{\left (l-1 \right )}$, the kernel height is $H^{\left (l-1 \right )}$, and the numbers of input and output channels are $C^{\left (l-1 \right )}$ and $C^{\left (l \right )}$, respectively.
Fig.~\ref{fig:weight_vectors} illustrates the weight vectors in fully connected and convolution layers. 
Note that the number of channels equals the number of nodes in the fully connected layer, which corresponds to the case of $W^{\left ( l-1 \right )} = H^{\left ( l-1 \right )} = 1$.

Let $\eta(\cdot)$ be an activation function. An input vector to a layer $l$ is then computed as ${\bf x}^{\left ( l \right )} = \eta \left ( {{\bf W}^{\left ( l \right )}}^\top  {\bf x}^{\left ( l-1 \right )} + {\bf b}^{\left( l \right)} \right )$ for the fully connected layer.
${\bf x}^{\left ( l \right )}$ should be defined at each spatial location for the convolution layer, but we use the notation of the fully connected layer for simplicity.
The parameters ${\bf \Theta}$ are learned by the optimization problem
\begin{align}
	\min_{\bf \Theta} \frac{1}{N} \sum_{n=1}^N {\mathcal L} \left( {\bf u}_n, {\bf v}_n, {\bf \Theta} \right ) + {\mathcal R} \left( {\bf \Theta} \right ), 
\end{align}
where ${\mathcal L}(\cdot)$ and ${\mathcal R}(\cdot)$ represent a loss function and a regularization function, respectively. ${\bf u} \in \mathbb{R}^{D_{in}}$ and ${\bf v} \in \mathbb{R}^{D_{out}}$ are an input and a target vector, respectively, for the network, 
and $N$ is the number of samples in the training set. 
Instead of using all samples in each timestep of the optimization, we utilize a mini-batch of size $M$, in which samples are drawn independently and uniformly from the training set. 

%Model Reduction
\subsection{Model Reduction}\label{subsec:model_reduction}
To reduce the model size, we utilize group sparsity by pruning weights under a threshold, that is, we prune column vectors ${\bf w}_k^{\left ( l \right )} \left ( k = 1,\dots,C^{\left ( l\right )} \right )$ in ${\bf W}^{\left ( l \right)}$, satisfying $ \Vert {\bf w}_k^{\left ( l \right )} \Vert _2 < \xi$ for a small positive constant $\xi$.
We induce sparsity by imposing the $L_2$ regularization on each weight vector. We therefore define the regularization function as
\begin{align}
	{\mathcal R} \left( {\bf \Theta} \right ) = \frac{\lambda}{2} \sum_{l=2}^L \Vert {\bf W}^{\left( l \right)} \Vert_F^2 \hspace{5mm} \left( \lambda > 0 \right ),
\end{align}
where $\lambda$ is a regularization parameter, and $\Vert {\cdot} \Vert_F$ indicates the \textit{Frobenius} norm.
We use ReLU \cite{Glorot_2011}, given as $\eta(x) = \max(x, 0)$, as the activation function for each non-output layer, and optimize the objective function by using Adam \cite{Kingma_2015}. 
Its update rule for a parameter $\theta$ is given by
\begin{align}\label{eq:adam}
	\left \{
		\begin{array}{ll}
		m_t = \beta_1 m_{t-1} + \left( 1 - \beta_1 \right)g_t																				&	\left( 0 < \beta_1 < 1 \right ) \\
		v_t = \beta_2 v_{t-1} + \left( 1 - \beta_2 \right)g_t^2																				&	\left( 0 < \beta_2 < 1 \right ) \\
		\theta_t = \theta_{t-1} - \alpha \frac{m_t / \left( 1 - \beta_1^t \right )}{\sqrt{v_t / \left( 1 - \beta_2^t \right) + \epsilon} }	&	\left( \epsilon > 0 \right )
		\end{array},
	\right.
\end{align}
where $g_t$ is the gradient at timestep $t$; $\beta_1$, $\beta_2$, and $\epsilon$ are positive constants.

The flow of our method is shown in Algorithm \ref{alg:learning} as pseudocode.
After training the network, the size is reduced by eliminating all weight vectors having $L_2$ norm smaller than a threshold $\xi$.
Since the weights converge to near the origin, it is easy to select the threshold. 
Our method can also effectively reduce the model size by directly eliminating channels or nodes.

%%%%%%%%%%%%%%%%%%%%%%%

%%%%%%%%%%%%%%%%%%%%%%%
%Theoretical Analysis
\section{Theoretical Analysis}\label{sec:theoretical_analysis}
In this section, we mathematically describe the mechanism of the sparsity and analyze convergence under certain conditions.

%preliminaries
\subsection{Preliminaries}\label{subsec:preliminaries}
The gradient with respect to ${\bf w}_k^{\left( l \right)}$, the $k$th weight vector in layer $l$, is given by
\begin{align}
	{\bf g}_k^{\left( l \right)} 
	= \frac{1}{M} \sum_{i=1}^M
	\frac{\partial {\mathcal L} \left( {\bf u}_i, {\bf v}_i , {\bf \Theta} \right)}
	{\partial x_{ik}^{\left( l\right)}}
	\frac{\partial x_{ik}^{\left( l\right)}}
	{\partial {\bf w}_k^{\left( l \right)}}
	+ \lambda {\bf w}_k^{\left( l \right)}.
\end{align}
If the activation function is ReLU, 
\begin{align}
	\frac{\partial x_{ik}^{\left( l\right)}}
	{\partial {\bf w}_k^{\left( l \right)}}
	=
	\left \{
		\begin{array}{ll}
		{\bf x}_i^{\left( l-1 \right)}	&	({{\bf w}_k^{\left( l \right)}}^\top {\bf x}_i^{\left( l-1 \right)} + b_k^{\left( l \right) } > 0) \\
		{\bf 0}							&	({{\bf w}_k^{\left( l \right)}}^\top {\bf x}_i^{\left( l-1 \right)} + b_k^{\left( l \right) } \leq 0)
		\end{array},
	\right.	
\end{align}
then the gradient becomes
\begin{align}
	{\bf g}_k^{\left( l \right)} 
	= \frac{1}{M} \sum_{ j:{{\bf w}_k^{\left( l \right)}}^\top {\bf x}_j^{\left( l-1 \right)} + b_k^{\left( l \right) } > 0 }
	\frac{\partial {\mathcal L} \left( {\bf u}_j, {\bf v}_j , {\bf \Theta} \right)}
	{\partial x_{jk}^{\left( l\right)}}
	{\bf x}_j^{\left( l-1 \right)} + \lambda {\bf w}_k^{\left( l \right)}.
\end{align}
Thus, if there are only a few samples satisfying ${{\bf w}_k^{\left( l \right)}}^\top {\bf x}_j^{\left( l-1 \right)} + b_k^{\left( l \right) } > 0$ (\ie, activated samples), 
${\bf g}_k^{\left( l \right) } \approx \lambda {\bf w}_k^{\left( l \right)}$, and
the vanilla-SGD is used as the optimizer with a step size $\alpha$, then the weight is updated as ${\bf w}_k^{\left( l \right)} \gets {\bf w}_k^{\left( l \right)} - \alpha \lambda {\bf w}_k^{\left( l \right)} = \left ( 1 - \alpha \lambda \right ) {\bf w}_k^{\left( l \right)}$, 
which means the weight decays to zero if $0 < \alpha \lambda < 1$.
From the observation that 50 -- 90\% of ReLUs are not activated~\cite{Glorot_2011}\cite{Shi_2017}, 
we assume the gradient of the regularization should be dominant for weights connected to such less-activated ReLUs, which induces the convergence to zero.
As shown in Fig.~\ref{fig:adam_sparsity}, however, this behavior does not appear with mSGD.
We believe this is due to the difference in convergence rate between the optimizers.

\begin{figure}[tb] 
	\centerline{\includegraphics[width = 85mm]{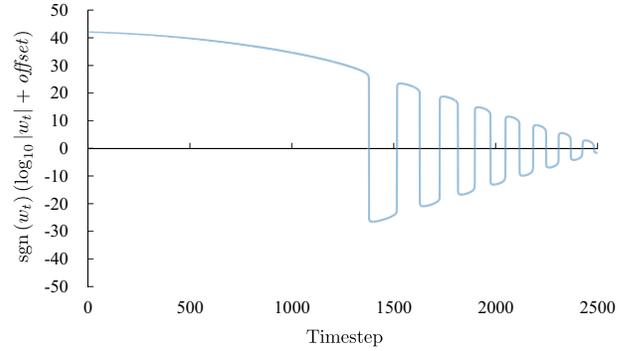}}
	\caption{Convergence behavior of $w_t$ under Adam. The zero point in the vertical axis corresponds to near the origin $(10^{-40})$.}
	\label{fig:adam_simulation}
\end{figure}

We conducted a preliminary experiment with the Adam optimizer under the condition that ${\bf g}_k^{\left( l \right) } \approx \lambda {\bf w}_k^{\left( l \right)}$, 
and found that after the weight decays to a certain level, it oscillates near the origin, as shown in Fig.~\ref{fig:adam_simulation}.
For example, a weight decreased from an initial value of 0.2 and then began to oscillate around $10^{-10}$. 
This oscillation is not observed with mSGD.
In Theorem~\ref{thm:adam} in the following subsection, we give the decay rate for Adam under the pre-oscillation condition.

\begin{algorithm}[tb]
	\caption{Learning and reducing network}
	\label{alg:learning}
	\begin{algorithmic}[1]
	\REQUIRE Network with ReLU activations and learnable parameters ${\bf \Theta} = \left \{ {\bf W}^{\left( l \right)}, {\bf b}^{\left( l \right)} \right \}_{l=2}^L$
	\REQUIRE Stochastic objective function with $L_2$ regularization
	\REQUIRE Initial parameters ${\bf \Theta}_0$, an $L_2$-regularization parameter $\lambda > 0$, and a norm threshold parameter $\xi > 0$
		\STATE{$t \leftarrow 0$ (Initialize timestep)}
		\WHILE{${\bf \Theta}_t$ not converged}
			\STATE{$t \leftarrow t + 1$}
			\STATE{Compute ${\bf \Theta}_t$ with Adam}
		\ENDWHILE
		\FOR{$l = 2, \dots, L$}
			\STATE{Eliminate $ \left \{ {\bf w}_k^{\left( l \right)} \relmiddle| \Vert {\bf w}_k^{\left( l \right)} \Vert_2 < \xi,\ k = 1, \dots, C^{\left( l \right)} \right \} $ \\ from ${\bf \Theta}_t$}
		\ENDFOR
		\RETURN{learned network with reduced parameters ${\bf \Theta}_t$}
	\end{algorithmic}
\end{algorithm}

%Convergence Analysis
\subsection{Convergence Analysis}\label{subsec:convergence_analysis}
Here, for simplicity of notation, we omit the indices $k$ and $l$ and use the subscript $t$ to indicate the timestep, then use a scalar weight denoted by $w_t$.
We suppose a non-activation situation in which the gradient of the loss is zero; thus, $g_t = \lambda w_t$. 
Detailed proofs of the following theorems are given in the appendices.
\theoremstyle{plain}
\newtheorem{thm}{Theorem}[section]
\newtheorem{lem}[thm]{Proposition}
\newtheorem{prop}[thm]{Proposition}
\begin{prop}\label{prop:msgd}
Let $\mu = 1 + \alpha \lambda - q \sqrt{\alpha \lambda}$ for $q > 2$. Under conditions $(1-\alpha \lambda + \mu)^2 - 4\mu \geq 0$ and $0 < \lambda \alpha < 1$, 
$w_t$ decays at rate $O\left( \left(1 - \frac{q - \sqrt{q^2 - 4}}{2} \sqrt{\alpha \lambda}  \right)^t \right)$ for mSGD. \\
\end{prop}
\noindent Therefore, if $\frac{q - \sqrt{q^2 - 4}}{2} \sqrt{\alpha \lambda} \geq \alpha \lambda$, then 
we obtain a convergence rate faster than the $O\left( \left(1 - \alpha \lambda  \right)^t \right)$ given by SGD.
In particular, we obtain a convergence rate similar to Nesterov's acceleration \cite{Nesterov_2004}. 

\begin{thm}\label{thm:adam}
Without loss of generality, we may suppose $w_0 > 0$. Then, as long as $t$ satisfies that $w_\tau > 0$ for $1 \leq \tau \leq t$,
$w_t$ decays as $w_t = O\left( \exp \left( -2^t \right) \right )$ for Adam.
\end{thm}
\noindent The above convergence rate is doubly exponential. Therefore, to achieve 
$
w_{t} \leq \delta,
$
we require only 
$
t= \Omega(\log\log(1/\delta))
$
steps. We recall that SGD and mSGD require
$
t = \Omega(\log(1/\delta))
$
steps to achieve $w_{t} \leq \delta$. 
Thus, the solution with Adam becomes smaller than sufficiently small $\delta$ more rapidly than with SGD and mSGD. 
We believe this rapid decay before oscillation contributes to the sparsity with Adam.

Although our theorem for Adam does not guarantee convergence to the origin, we have the following proposition.
\begin{prop}
If $w_t$ converges to $w_{*}$ as $t \to \infty$,
then the limit point must satisfy $w_{*} = 0$.  \\
\end{prop}
\noindent Substituting a condition: $g_t = \lambda w_t$ into Eq.~(\ref{eq:adam}), the step size of Adam for a sufficiently large $t$ can be represented as \\
\begin{align*}
|w_t - w_{t-1}| = \left|- \alpha \frac{1 - \beta_1}{\sqrt{\left( 1 - \beta_2 \right)}} \frac{\sum_{i=0}^{t-1}\beta_1^{t-1-i}w_i}{\sqrt{\sum_{i=0}^{t-1}\beta_2^{t-1-i}w_i^2 + \epsilon}} \right|.
\end{align*}
If $w_t$ converges to $w_{*}$, 
then as $\sum_{\tau=t}^\infty \beta_1^\tau \to 0$, 
$\sum_{\tau=t}^\infty \beta_2^\tau \to 0$ for $t\to \infty$,
it holds that 
letting $\tilde{\epsilon} = \frac{\epsilon}{\sum_{i=0}^t \beta_2^{i}}$, the step size becomes
\footnotesize
\begin{align*}
|w_{t+1} - w_{t}| 
= &  \left|- \alpha \frac{\left(1 - \beta_1 \right)\sum_{i=0}^t
\beta_1^{i}}{\sqrt{\left( 1 - \beta_2 \right)\sum_{i=0}^t\beta_2^{i}}} \frac{w_{*} + o(1)}{\sqrt{\left(w_{*} \right)^2 + \frac{\epsilon}{\sum_{i=0}^t \beta_2^{i}}+ o(1)}} \right| \\
= &  \left|- \alpha \frac{\left(1 - \beta_1^t \right)}{\sqrt{\left( 1 - \beta_2^t \right)}} \frac{w_{*}}{\sqrt{\left(w_{*} \right)^2 + \tilde{\epsilon}}} 
+ o(1)
\right| \\
\stackrel{t \to \infty}{\longrightarrow}  & \left|- \alpha \frac{w_{*}}{\sqrt{\left(w_{*} \right)^2 + \tilde{\epsilon}}
} \right|.
\end{align*}
\normalsize
The above must be a \textit{Cauchy} sequence if it converges; that is, the step size must be 0, which is satisfied with $w_*=0$. \\
%%%%%%%%%%%%%%%%%%%%%%%

%%%%%%%%%%%%%%%%%%%%%%%
%Related Work
\section{Related Work}\label{sec:related_work}
In this section, we briefly review related works, including analyses on training DNNs from the viewpoints of activation functions and optimizers, and methods of model reduction.
Group sparsity as discussed in this paper relates to the \textit{vanishing gradient problem} \cite{Bengio_1994}, which is generally caused by saturation nonlinearities such as $\mathrm{sigmoid}$ and $\tanh$.
These functions have certain regimes in which the gradient vanishes, and ReLU also possesses such a regime (specifically, $x < 0$). 
Li \etal~\cite{JLi_2017} showed that unsupervised pretraining encourages sparse activations with $\mathrm{sigmoid}$ and ReLU in the resulting DNNs.
Though it has been known that some units never activate across the entire training dataset, the so-called \textit{dying-ReLU} \cite{Li_2015}, detailed analysis of this has not been reported. 
The remedy for this problem is leaky-ReLU \cite{Maas_2013} given by 
\begin{align}
	\eta(x)
	 = 
	\left \{
		\begin{array}{ll}
		x		&	\left(x > 0 \right) \\
		\rho x	&	\left(x \leq 0 \right) 
		\end{array},
	\right.
\end{align}
which parameterizes the negative slope $(\rho)$ to prevent its gradients from vanishing in the negative regime.
In the next section, we compare various activation functions, including leaky-ReLU and saturation nonlinearities, in terms of sparsity. 

SGD is a common optimizer for training DNNs. 
It is a simple method but achieves good generalization performance by careful tuning of step-size and acceleration via the momentum term \cite{Qian_1999}.
In contrast, adaptive gradient methods, including Adagrad \cite{Duchi_2011}, RMSprop \cite{Tieleman_2012}, Adam \cite{Kingma_2015}, and AdaDelta \cite{Zeiler_2012}, commonly attain faster convergence than the SGD by automatically adjusting the step size.
Compared with the SGD, however, their generalization performances are worse in many cases. Theoretical analysis has been conducted, and improvements have been reported \cite{Wilson_2017,Reddi_2018,Loshchilov_2017}.
Wilson \etal~\cite{Wilson_2017} proved that the adaptive methods do not converge to an optimum in a simple convex optimization problem.
Reddi \etal~\cite{Reddi_2018} proved that the non-convergence of Adam is due to the step-size scaling according to the moving average of the squared gradient, and proposed an improved method, AMSGRAD.
Loshchilov \etal~\cite{Loshchilov_2017} proposed AdamW, which decouples the weight-decay term from the update rule of Adam, and showed it achieves performance similar to that of mSGD.
Despite these analyses of Adam, it has not been reported that Adam induces implicit weight sparsity in training DNNs.
In the next section, we compare Adam against various optimizers including mSGD, adaptive gradient methods, and the improved methods of Adam \cite{Reddi_2018,Loshchilov_2017}. 

There are two broad approaches to model reduction: pruning redundant weights after learning and learning with sparsity-inducing regularizers on weights.
In an example of the former approach, Li \etal~\cite{Li_2017} exploit the $L_1$ norm of weight vectors to prune less important channels after learning the DNN, and Polyak \etal~\cite{Polyak_2015} do the same based on the statistics of ReLU activations. 
However, selecting the pruning threshold is difficult, and fine-tuning after pruning is necessary for recovering the accuracy. 
In our method, since the weights converge to near the origin, it is easy to select the threshold, which only need to be smalls, such as $10^{-15}$, and the fine-tuning is optional.
In the latter approach, Wen \etal~\cite{Wen_2016}, Scardapane \etal~\cite{Scardapane_2017}, and Yoon \etal~\cite{Yoon_2017} minimize the $L_{2,1}$ norm of weight vectors with the objective function so that channel-level group sparsity is obtained.
Our method yields sparsity based on the $L_2$ norm, which is a smooth function, making it easy to optimize. We show the effectiveness of our method relative to the other methods \cite{Wen_2016}\cite{Scardapane_2017} below. 
%%%%%%%%%%%%%%%%%%%%%%%

%%%%%%%%%%%%%%%%%%%%%%%
%Experiments
\section{Experiments}\label{sec:experiments}

\begin{table}[t]
\begin{center}
	\caption{Baseline setup of experiments on MNIST dataset.}
	\begin{tabular}{l|l} \hline
	Preprocess				&	divide each pixel value by 255													\\
	Data augmentation		&	No																				\\
	Batch size / \#epochs	&	64 / 100																		\\
	Learning rate schedule	&	multiplied by 0.5 every 25 epochs												\\
	$L_2$-norm threshold	&	$\xi = 1.0 \times 10^{-15}$														\\
	$L_2$ regularization	&	Yes ($\lambda = 5.0 \times 10^{-4}$)											\\
	\# hidden layers		&	1																				\\
	\# nodes per layer		&	1000																			\\
	Activation function		&	ReLU																			\\
	Batch normalization		&	Yes (before activation)															\\
	Initializer				&	Xavier \cite{Glorot_2010} for weights, 0 for biases								\\
	Optimizer				&	Adam																			\\
							&	{\scriptsize($\alpha = 0.001, \beta_1=0.9, \beta_2=0.999, \epsilon=10^{-8}$)}	\\ \hline
	\end{tabular}
	\label{tbl:setup_mnist}
\end{center}
\end{table}

\begin{table}[t]
\begin{center}
	\caption{Comparisons with different optimizers, activation functions, and other components on MNIST dataset.}
	\begin{tabular}{l|cccc} \hline
	Optimizer				&	Adam (baseline)	&	mSGD		&	adagrad	&	RMSprop				\\ \hline
	Acc. [\%]				&	98.43			&	98.35		&	98.31	&	98.31				\\
	Sparsity [\%]			&	70.00			&	0.00		&	0.00	&	65.20				\\ \hline
	\hline
	Activation				&	ReLU (baseline)	&	$\tanh$		&	ELU		&	$\mathrm{sigmoid}$	\\ \hline
	Acc. [\%]				&	98.43			&	98.39		&	98.27	&	98.42				\\
	Sparsity [\%]			&	70.00			&	1.70		&	77.00	&	0.00				\\ \hline
	\hline
	Others					&	baseline		&	He init.	&	w/o BN	&	w/o $L_2$ reg.		\\ \hline
	Acc. [\%]				&	98.43			&	98.42		&	98.26	&	98.45				\\
	Sparsity [\%]			&	70.00			&	68.70		&	82.10	&	0.00				\\ \hline
	\end{tabular}
	\label{tbl:opt_mnist}
\end{center}
\end{table}

\begin{table*}[t]
\begin{center}
	\caption{Comparisons with different numbers of hidden nodes on MNIST dataset.}
	\begin{tabular}{l|cccccc} \hline
	\# of nodes				&	10		&	50		&	100		&	500		&	1000 (baseline)	&	2000	\\ \hline
	Acc. [\%]				&	93.92	&	97.46	&	98.17	&	98.48	&	98.43			&	98.45	\\
	\# of remaining weights	&	10		&	50		&	100		&	291		&	300				&	288		\\ \hline
	\end{tabular}
	\label{tbl:nodes_mnist}
\end{center}
\end{table*}

\begin{table*}[t]
\begin{center}
	\caption{Comparisons with different numbers of hidden layers on MNIST dataset.}
	\begin{tabular}{l|ccccc} \hline
	\# of hidden layers								&	1 (baseline)	&	2		&	3			&	4				&	5					\\ \hline
	Acc. [\%]										&	98.43			&	98.73	&	98.78		&	98.88			&	98.68				\\
	\# of remaining weights in each hidden layer	&	300				&	128-227	&	153-75-186	&	177-78-76-160	&	168-87-73-75-160	\\ \hline
	\end{tabular}
	\label{tbl:layers_mnist}
\end{center}
\end{table*}

\begin{figure}[t]
	\centerline{\includegraphics[width = 88mm]{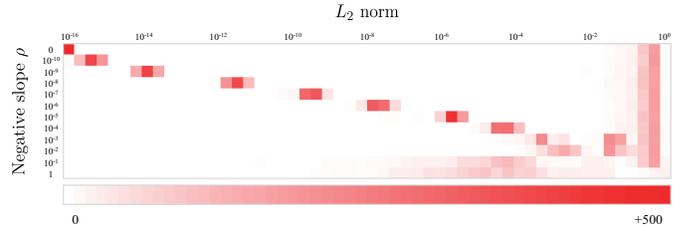}}
	\caption{Distributions of $L_2$ norm per weight vector for the negative slope of leaky-ReLU.}
	\label{fig:r_relu}
\end{figure}

\begin{figure}[t]
	\centerline{\includegraphics[width = 85mm]{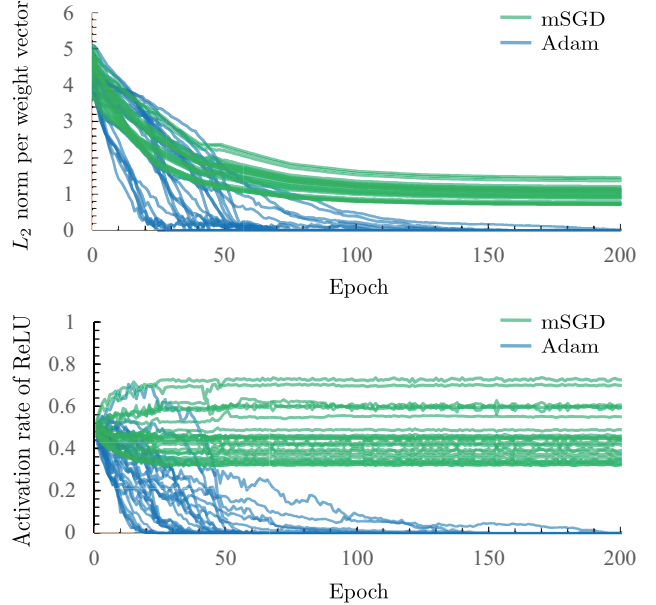}}
	\caption{
		Progress of $L_2$ norm (upper) and activation rate (lower) with 23 convolution filters in first layer of a VGG-style network. 
		The network was trained with Adam and mSGD on the CIFAR-10 dataset.}
	\label{fig:norm_activation}
\end{figure}

In this section, we demonstrate the sparsity under various training setups and compare our model reduction method with others that use sparsity-inducing regularizers. The experiments were on image classification tasks using MNIST \cite{LeCun_1998} and CIFAR-10 datasets \cite{Krizhevsky_2009}. 
In all experiments on both datasets, we used softmax function in output layer and cross-entropy loss for training networks. 

%Demonstration of the sparsity
\subsection{Demonstration of Sparsity}\label{subsec:demo_sparse}
On the MNIST dataset, we used fully connected networks with the experimental setup summarized in Table \ref{tbl:setup_mnist} as the baseline, and evaluated different activation functions, optimizers, and other components.
Table \ref{tbl:opt_mnist} shows the validation accuracy and sparsity of the weight vectors produced by the different setups. 
There is not much difference in terms of accuracy, but the sparsity varies for each component.
Near the top of Table \ref{tbl:opt_mnist}, RMSprop yields sparsity as well as that of Adam.
RMSprop is basically a special case of Adam by setting $\beta_1=0$, and we see it has similar convergence. 
The middle of Table \ref{tbl:opt_mnist} shows the results for different activation functions. 
Sparsity is obtained by ReLU and by exponential linear units (ELU) \cite{Clevert_2016}.
Since its gradients mostly vanish in the negative regime except near the origin, we see it behaves like ReLU, that is, the gradient of the objective function is dominated by that of $L_2$ regularization, and it yields similar sparsity. 
While $\tanh$ obtains only limited sparsity, it indicates that saturation nonlinearities have the potential for sparsity. 
We also investigated the behavior of leaky-ReLU \cite{Maas_2013}. 
We plotted the distributions of $L_2$ norm with changing its negative slope $\rho$ as shown in Fig.~\ref{fig:r_relu}. 
In the case of $\rho$ = 0, which corresponds to ReLU, we can see there are two modes: the lower mode for (near-) zero weights and the other mode for active weights.
Observing that the lower mode shifts as the negative slope increases, we see that leaky-ReLU can alleviate the decay of weights by propagating gradients for negative inputs. 

Additionally, we evaluated the behavior with an initializer proposed by He \etal~\cite{He_2015} and with or without BN layer and $L_2$ regularization.
The bottom of Table \ref{tbl:opt_mnist} shows that the initializer and BN layer make no appreciable difference, but $L_2$ regularization is needed for sparsity, as expected. 
Next, we investigated the effects of the number of nodes in the hidden layer and the number of layers in the network.
Table \ref{tbl:nodes_mnist} shows the validation accuracy and number of retained weight vectors (\ie, nodes) after training for the different number of nodes. 
It can be observed that networks having a small number of nodes results in low accuracy and the weights of those are fully retained. 
We attribute this to such networks not being redundant, meaning that each node in the networks contributes to decreasing the loss (\ie, each node is well activated). In contrast, the numbers of retained weights (and accuracies) in networks with more than 500 nodes were almost identical. 
This result indicate that wide networks can be reduced to a certain narrow model without sacrificing accuracy. 
The results with different number of layers are shown in Table \ref{tbl:layers_mnist}, 
where each network was trained with 1000 nodes in each layer. 
The results indicate that the number of retained weights becomes small in the deeper layers (except for the last hidden layer). 

On the CIFAR-10 dataset, we trained a VGG-style covolutional neural network with almost the same setup as in \cite{github}. 
During a preprocessing step, each image was normalized to have mean 0 and standard deviation 1 over three channels, and horizontal flipping was applied as data augmentation. 
We again used the initializer proposed by He \etal~\cite{He_2015} for weights and applied dropout \cite{Srivastava_2014} to convolution and fully connected layers, as in~\cite{github}. 
With the same parameter of $\lambda = 5.0 \times 10^{-4}$, we compared against other optimizers: mSGD, AMSGRAD \cite{Reddi_2018}, and AdamW \cite{Loshchilov_2017}. 
An initial learning rate of 0.1 was used for mSGD and 0.0005 for the others.
The batch size, learning rate scheduling, and $L_2$-norm threshold for model reduction are shown in Table \ref{tbl:setup_mnist}.
The maximum validation accuracy during the training for 400 epochs and the ratio of reduced parameters at that time are reported in Table \ref{tbl:opt_cifar}.
We can see that parameters are not reduced by AMSGRAD and AdamW, that is, these optimizers do not induce sparsity.
Instead of using the moving average of the squared gradient, AMSGRAD uses its maximum until each timestep. 
Thus, AMSGRAD does not induce sparsity because the step-size tends to decrease as the training progresses, and so weights decay more slowly. 
For AdamW, the lack of sparsity is because the rate of decay in the weights is similar to that of SGD since the weight-decay term is decoupled from the step-size computation of Adam.
We can see that Adam achieves accuracy similar to that of AMSGRAD but reduces more than 50\% of parameters.
However, there is still a gap in accuracy between Adam and mSGD.
We suspect that this gap may result from the implicit weight sparsity of Adam. 

Next, we observed relationships between activation rate of channels (ReLUs) and $L_2$ norm of their weight vectors during the training. 
We again trained a VGG-style network by Adam and mSGD with the same setups as above.
We picked up 23 of 64 filters in first convolution layer, which resulted in convergence to zero with Adam.
The progress of their activation rate and $L_2$ norm with both optimizers is illustrated in Fig.~\ref{fig:norm_activation}, respectively. 
Activation rate of a filter was computed as follows: 
counted nonzero pixels in a feature map after the ReLU layer; 
divided it by the feature map size; and 
computed it for all training images, then averaged it.
In Fig.~\ref{fig:norm_activation}, we can see that $L_2$ norm and activation rate with Adam jointly decrease to zero while those with mSGD converge to some nonzero values as the training progress. 
This observation indicates that activation rate of ReLU decreases in conjunction with $L_2$ norm of its input weight vector. 
Thus, we see that the weight decays rapidly with Adam even when ReLU is not completely inactive (\ie, a condition: $g_t = \lambda w_t$ is not exactly satisfied). 

%Comparisons on model reduction
\begin{table}[t]
\begin{center}
	\caption{Comparisons with different optimizers on CIFAR-10 dataset.}
	\begin{tabular}{l|cc} \hline
	Optimizer					&	Acc. [\%]	&	Reduced [\%]	\\ \hline
	mSGD (as in \cite{github})	&	92.45		&	-				\\
	mSGD (our implementation)	&	93.13		&	0.00			\\
	AMSGRAD 					&	92.64		&	0.00			\\
	AdamW 						&	91.97		&	0.00			\\
	Adam 						&	92.61		&	53.23			\\ \hline
	\end{tabular}
	\label{tbl:opt_cifar}
\end{center}
\end{table}

\begin{table}[t]
\begin{center}
	\caption{Comparisons with other model reduction method on MNIST dataset.}
	\begin{tabular}{l|c|c|c} \hline
	\hspace{-1.5mm}Method									\hspace{-1.5mm}	&	\hspace{-1.5mm}Error [\%]	\hspace{-1.5mm}	&	\hspace{-1.5mm}Reduced [\%]	\hspace{-1.5mm}	&	\hspace{-1.5mm}\#Neurons 		\hspace{-1.5mm}	\\ \hline
	\hspace{-1.5mm}Baseline (cited from \cite{Wen_2016})	\hspace{-1.5mm}	&	\hspace{-1.5mm}1.43			\hspace{-1.5mm}	&	\hspace{-1.5mm}- 			\hspace{-1.5mm}	&	\hspace{-1.5mm}784-500-300-10	\hspace{-1.5mm}	\\
	\hspace{-1.5mm}Wen \etal\ (cited from \cite{Wen_2016})	\hspace{-1.5mm}	&	\hspace{-1.5mm}1.53			\hspace{-1.5mm}	&	\hspace{-1.5mm}83.5			\hspace{-1.5mm}	&	\hspace{-1.5mm}434-174-78-10 	\hspace{-1.5mm}	\\
	\hspace{-1.5mm}ours										\hspace{-1.5mm}	&	\hspace{-1.5mm}1.45			\hspace{-1.5mm}	&	\hspace{-1.5mm}83.7			\hspace{-1.5mm}	&	\hspace{-1.5mm}784-91-174-10	\hspace{-1.5mm}	\\ \hline
	\end{tabular}
	\label{tbl:comp_mnist}
\end{center}
\end{table}

\begin{figure}[t] 
	\centerline{\includegraphics[width = 90mm]{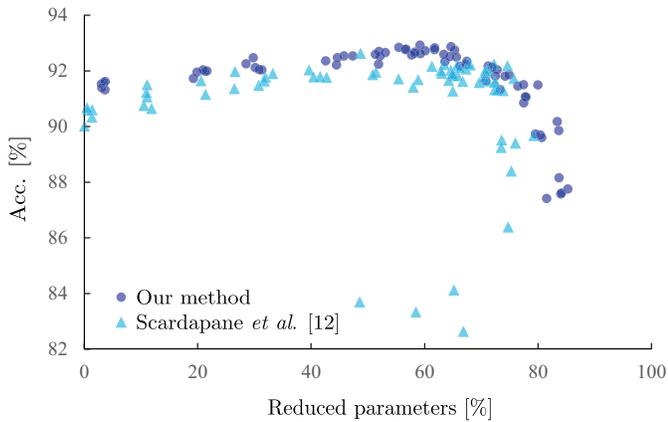}}
	\caption{
	Trade-off between accuracy and ratio of reduced parameters on CIFAR-10 dataset. 
	The following regularization parameters were used: 
	$\{1.0\times10^{-5},\>5.0\times10^{-5},\>1.0\times10^{-4},\>2.5\times10^{-4},\>5.0\times10^{-4},\>7.5\times10^{-4},\>0.001,				\>0.0015,			\>0.0025,				\>0.005,			\>0.0075,				\>0.01,				\>0.025,			\>0.05\}$ for our method and 
	$\{1.0\times10^{-6},\>2.0\times10^{-6},\>3.0\times10^{-6},\>4.0\times10^{-6},\>5.0\times10^{-6},\>7.0\times10^{-6},\>9.0\times10^{-6},	\>1.0\times10^{-5},	\>1.25\times10^{-5},	\>1.5\times10^{-5},	\>1.75\times10^{-5},	\>2.0\times10^{-5},	\>3.0\times10^{-5},	\>4.0\times10^{-5}\}$ for the other method~\cite{Scardapane_2017}.
	}
	\label{fig:trade_off_cifar}
\end{figure}

\subsection{Comparisons on Model Reduction}\label{subsec:comp_on_reduction}
We conducted comparisons with other model reduction methods in which the group sparsity of weights is explicitly induced by the regularizer. 
On the MNIST dataset, we compared with a method proposed by Wen \etal~\cite{Wen_2016}. 
We used the same network model as in \cite{Wen_2016}, namely, a 4-layer fully connected network without BN layer, having 500 and 300 nodes in its hidden layers.
We trained it with He's initializer \cite{He_2015} and the same setup as in Table \ref{tbl:setup_mnist}.
Our method is slightly better than \cite{Wen_2016} in accuracy while comparable in reduction rate, as shown in Table \ref{tbl:comp_mnist}. 

Moreover, we compared with a method proposed by Scardapane \etal~\cite{Scardapane_2017} on the CIFAR-10 dataset, using the same model and setup as the experiment in a previous subsection.
We implemented their method, in which the model was trained by mSGD with initial learning rate of 0.1. 
The threshold of model reduction was set as $1.0\times10^{-6}$ for the other method.
For investigating the trade-off between accuracy and reduction rate, we used several parameters of the regularization $(\lambda)$.
The network was trained 5 times for each $\lambda$ with different initial weights, and the accuracy and ratio of reduced parameters were plotted in Fig.~\ref{fig:trade_off_cifar}.
It can be seen that our method basically outperforms the other method in accuracy while achieving the same reduction rate. 
The other method sometimes achieved low accuracy, and ours was more stable.
Therefore, our method can efficiently control the model size by calibrating a parameter of $L_2$ regularization.
%%%%%%%%%%%%%%%%%%%%%%%

%%%%%%%%%%%%%%%%%%%%%%%
%Conclusions
\section{Conclusions}\label{sec:conclusions}
In this paper, we have analyzed the implicit weight sparsity induced by Adam, ReLU, and $L_2$ regularization in both theoretical and experimental aspects.
Assuming that weight decay by $L_2$ regularization becomes dominant under the existence of less-activated ReLUs, 
we have mathmatically described that Adam requires $\Omega(\log\log(1/\delta))$ steps to achieve a sufficiently small weight of $\delta$, which is faster than the $\Omega(\log(1/\delta))$ of mSGD. 
We believe this difference leads to sparsity of weights with Adam. 
Additionally, we proposed a method for model reduction by simply eliminating the zero weights after training the DNN. 
In experiments on MNIST and CIFAR-10 datasets, we demonstrated the sparsity with various training setups and 
found that other activation functions and optimizers having properties similar to ReLU and Adam, respectively, also achieve sparsity.
Finally, we confirmed that our method can efficiently reduce the model size and exhibits favorable performance relative to that of other methods that use a sparsity-inducing regularizer. 
%%%%%%%%%%%%%%%%%%%%%%%

%%%%%%%%%%%%%%%%%%%%%%%
%Acknowledgment
\section*{Acknowledgment}
\addcontentsline{toc}{section}{Acknowledgment}
TS was partially supported by MEXT Kakenhi (25730013, 25120012, 26280009, 15H05707 and 18H03201), Japan Digital Design, JST-PRESTO and JST-CREST.
%%%%%%%%%%%%%%%%%%%%%%%

%%%%%%%%%%%%%%%%%%%%%%%
%References
\bibliography{references}		% bibliography data in references.bib
\bibliographystyle{IEEEtran}	% makes bibtex using IEEEtran.bst
%%%%%%%%%%%%%%%%%%%%%%%

%%%%%%%%%%%%%%%%%%%%%%%
%Appendices
\section*{Appendices}
\addcontentsline{toc}{section}{Appendices}
\subsection{Convergence of mSGD (Proposition~\ref{prop:msgd})}
The update rule of mSGD for a weight $w_t$ is given by
\begin{align*}
	\left \{
		\begin{array}{ll}
		v_{t+1} = \mu v_{t} - \alpha \lambda w_t	&	\left( 0 < \mu < 1 \right ) \\
		w_{t+1} = w_t + v_{t+1}
		\end{array}.
	\right.
\end{align*}
The condition $(1-\alpha \lambda + \mu)^2 - 4\mu \geq 0$ implies that
\begin{align*}
& \mu^2 + 2(1-\alpha \lambda -2 ) \mu + (1 - \alpha \lambda)^2  \geq 0 \\
\Rightarrow &
  |\mu -( 1+\lambda \alpha)| \geq 2 \sqrt{\alpha \lambda}.
\end{align*}
Since $0 < \mu < 1$, $\mu$ must satisfy $\mu \leq 1 + \lambda \alpha - 2 \sqrt{\alpha \lambda}$
under the assumption $0 < \lambda \alpha < 1$.
Here, letting $\mu = 1 + \alpha \lambda - q \sqrt{\alpha \lambda}$ for $q > 2$
and defining 
$\beta = \frac{2 - q\sqrt{\alpha \lambda} + \sqrt{(q^2 - 4)\alpha \lambda}}{2}$ and 
$\gamma = \frac{2 - q\sqrt{\alpha \lambda} - \sqrt{(q^2 - 4)\alpha \lambda}}{2}$,
we have that 
\begin{align*}
w_t & = \left\{ (1 - \gamma) (\beta^t + \beta^{t-1} \gamma + \dots + \beta \gamma^{t-1}) + \gamma^t \right\} w_0  \\
&  =\left\{  (1 - \gamma) \beta^{t} \frac{1 -( \gamma/\beta)^t}{1 - \gamma/\beta} + \gamma^t \right\} w_0  \\
&  = \left\{ \frac{1 - \gamma}{\beta - \gamma} \beta^{t+1} [1 -( \gamma/\beta)^t] + \gamma^t \right\} w_0.
\end{align*}
The right-hand side can be evaluated as 
\begin{align*}
& \frac{1 - \gamma}{\beta - \gamma} \beta^{t+1} [1 -( \gamma/\beta)^t] + \gamma^t  
	\leq  \left[ \frac{(q + \sqrt{q^2 -4}) }{2 \sqrt{q^2 -4}} + 1 \right]\beta^t \\
&   = \left[ \frac{(q + \sqrt{q^2 -4}) }{2 \sqrt{q^2 -4}} + 1 \right]
    	\left(1 - \frac{q - \sqrt{q^2 - 4}}{2} \sqrt{\alpha \lambda}  \right)^t \\
&   = O\left( \left(1 - \frac{q - \sqrt{q^2 - 4}}{2} \sqrt{\alpha \lambda}  \right)^t \right). \\
\end{align*}

\subsection{Convergence of Adam (Theorem~\ref{thm:adam})}

Let $\nu = \frac{\alpha(1-\beta_1)}{\sqrt{1-\beta_2}}$. Then, by the update rule, $w_t$ is recursively given by 
$$
w_t = w_{t-1} - \nu \frac{ m_{t}}{\sqrt{v_t + \epsilon}},
$$
for $m_{t} = w_{t-1} + \beta_1 w_{t-2} + \dots + \beta_{1}^t w_0$ and $v_t =  w_{t-1}^2 + \beta_2 w_{t-2}^2 + \dots + \beta_{2}^t w_0^2$.

First, notice that if $\beta_1 \leq \beta_2$, then 
the difference between $w_t$ and $w_{t-1}$ is bounded by 
\begin{align*}
&	\nu \left| \frac{w_{t-1} + \beta_1 w_{t-2} + \dots + \beta_1^t w_0 }{\sqrt{v_t + \epsilon}} \right| \\
&	\leq \nu  \frac{ |w_{t-1}| + \beta_1 |w_{t-2}| + \dots + \beta_1^t |w_0| }{\sqrt{v_t + \epsilon}}  \\
& \leq \nu  \frac{ \sqrt{ w_{t-1}^2 + \beta_1 w_{t-2}^2 + \dots + \beta_1^{t} w_0^2} }{\sqrt{v_t + \epsilon}} (1+\beta_1 + \dots + \beta_1^{t}) \\
& \leq \nu \frac{\sqrt{v_t}}{\sqrt{v_t + \epsilon}} \frac{1}{1 - \beta_1} \leq \frac{\alpha}{\sqrt{1 - \beta_2}} =: \nu'.
\end{align*}
Hence, we see that, as long as $w_t > \nu'$, positivity of $w_{t+1}$ is ensured: $w_{t+1} > 0$.

Suppose that $w_0 > 0$ and $t$ is an integer such that, for $1 \leq t' \leq  t$, $w_{t'} > \delta \geq \nu'$.
In this setting, 
\begin{align*}
w_t & \leq w_{t-1} - \frac{\nu}{\sqrt{v_t + \epsilon}}  w_{t-1}  
 = w_{t-1} \left( 1 - \frac{\nu}{\sqrt{v_t + \epsilon}} \right).
\end{align*}
Therefore, we have $w_t < w_{t-1} < \dots < w_0$, and 
$$
v_{\tau +1} \leq w_0^2 (1 + \beta_2 + \beta_2^2 + \dots + \beta_2^\tau) \leq 
\frac{ w_0^2}{1 - \beta_2}.
$$
Here, letting $\xi = 1-\frac{\nu }{\sqrt{w_0^2/(1 - \beta_2) + \epsilon}}$, we have $w_{\tau} \leq \xi^{\tau} w_0~(1\leq \tau \leq t)$.
From this observation, 
we again evaluate $v_\tau$ as
\begin{align*}
v_{\tau+1} 
&	= w_\tau^2 + \beta_2 w_{\tau-1}^2 + \dots + \beta_2^\tau w_0^2 \\
&	\leq (\xi^\tau + \beta_2 \xi^{\tau-1} + \dots + \beta_2^\tau)w_0^2 \\
&	\leq (\tau + 1) \max\{\xi,\beta_2\}^\tau w_0^2.
\end{align*}
Now, let $\tau^*$ be the smallest $\tau$ such that $(\tau + 1) \max\{\xi,\beta_2\}^\tau
\leq 1/2$.
By this definition, 
if $k \tau^* \leq \tau \leq (k+1)\tau^*$, then
$$v_{\tau + 1} \leq (1/2)^k w_0^2.$$
Therefore, for $k \tau^* \leq \tau \leq (k+1)\tau^*$, 
$$
w_{\tau + 1} \leq w_{\tau}\left( 1 - \frac{\nu}{\sqrt{(1/2)^k w_0^2 + \epsilon}} \right).
$$
Now, supposing that $k$ satisfies $\epsilon \leq (1/2)^kw_0^2$ (otherwise, we have $w_\tau^2 \leq \epsilon$), the above inequality yields
\begin{align*}
w_{\tau  + 1}
& \leq \exp\left(  -  \tau^* \sum_{\kappa=1}^{k-1} \frac{\nu}{\sqrt{(1/2)^\kappa w_0^2 + \epsilon}}  \right) w_0 \\
& \leq \exp\left(  -  \tau^* \sum_{\kappa=1}^{k-1} \frac{\nu}{w_0} 2^{(\kappa-1)/2}\right) w_0 \\
& 
= \exp\left(  -  \tau^* \frac{\nu}{w_0} (2^{k-1}-1)  \right) w_0  \\
& \leq \exp\left(  -  \tau^* \frac{\nu}{w_0} (2^{(\tau/\tau^*)/2 -1}-1) \right) w_0,
\end{align*}
which is doubly exponential. Therefore, to achieve 
$
w_{\tau+1} \leq \delta,
$
we require only 
$$
\tau \geq \left(\frac{\tau^* }{\log(2)} + 1\right) \log \left[ w_0 \frac{\log \left( w_0/\delta  \right) }{\nu \tau^* } + 1\right]
= \Omega(\log\log(1/\delta)).
$$
%%%%%%%%%%%%%%%%%%%%%%%
\end{document}